\documentclass[11pt,letterpaper]{article}
\usepackage{ijcnlp2017}
\usepackage{times}
\usepackage{latexsym}
\usepackage{url}

\ijcnlpfinalcopy

\def\ijcnlppaperid{82}

\usepackage{yfonts}
\usepackage{graphicx} 
\usepackage{epsfig} 
\usepackage{subfigure}

\usepackage[linesnumbered,ruled,vlined]{algorithm2e}

\SetCommentSty{mycommfont}

\usepackage{verbatim}
\usepackage{etoolbox}
\usepackage{framed}
\usepackage{xcolor}
\usepackage{color}
\usepackage{amsmath}
\usepackage{amssymb}
\newcommand{\argmax}{\operatornamewithlimits{argmax}}

\usepackage{multirow}
\usepackage{arydshln}

\title{Neural Probabilistic Model for Non-projective MST Parsing}

\author{Xuezhe Ma \and Eduard Hovy \\
	    Language Technologies Institute \\
	    Carnegie Mellon University \\
	    Pittsburgh, PA 15213, USA \\
	    {\tt xuezhem@cs.cmu.edu, hovy@cmu.edu}}

\date{}

\begin{document}
\maketitle

\begin{abstract}
In this paper, we propose a probabilistic parsing model that defines a proper conditional
probability distribution over non-projective dependency trees for a given sentence, 
using neural representations as inputs. The neural network architecture is based on bi-directional 
LSTM-CNNs, which automatically benefits from both word- and character-level representations, 
by using a combination of bidirectional LSTMs and CNNs. On top of the neural network, we introduce a 
probabilistic structured layer, defining a conditional log-linear model over non-projective trees.
By exploiting Kirchhoff's Matrix-Tree Theorem~\cite{tutte1984graph}, the partition functions and 
marginals can be computed efficiently, leading to a straight-forward end-to-end model training procedure 
via back-propagation. We evaluate our model on 17 different datasets, across 14 different languages. 
Our parser achieves state-of-the-art parsing performance on nine datasets.
\end{abstract}

\section{Introduction}
\label{sec:intro}
Dependency parsing is one of the first stages in deep language understanding and has gained
interest in the natural language processing (NLP) community,  
due to its usefulness in a wide range of applications. 
Many NLP systems, such as machine translation~\cite{xie-mi-liu:2011:EMNLP}, 
entity coreference resolution~\cite{ng:2010:ACL,durrett-klein:2013:EMNLP,ma-hovy:2016:NAACL},
low-resource languages processing~\cite{mcdonald-EtAl:2013:Short,ma-xia:2014:P14-1}, 
 and word sense disambiguation~\cite{fauceglia-EtAl:2015:EVENTS}, are becoming more sophisticated, 
in part because of utilizing syntactic knowledge such as dependency parsing trees.

Dependency trees represent syntactic relationships through labeled directed edges between heads 
and their dependents (modifiers). In the past few years, several dependency parsing algorithms~\cite{Nivre:2004,McDonald:2005b,Koo:2010,ma-zhao:2012:POSTERS,ma2015probabilistic} have been proposed, whose 
high performance heavily rely on hand-crafted features and task-specific resources 
that are costly to develop, making dependency parsing models difficult to adapt to new languages 
or new domains.

Recently, non-linear neural networks, such as recurrent neural networks (RNNs) with long-short term memory (LSTM) 
and convolution neural networks (CNNs), with as input distributed word representations, 
also known as word embeddings, have been broadly applied, 
with great success, to NLP problems like part-of-speech (POS) 
tagging~\cite{collobert2011natural} and named entity recognition (NER)~\cite{TACL792}.
By utilizing distributed representations as inputs, these systems are capable of learning hidden information
representations directly from data instead of manually designing  hand-crafted features, yielding 
end-to-end models~\cite{ma-hovy:2016:P16-1}. Previous studies explored the applicability of neural
representations to traditional graph-based parsing models. Some work~\cite{TACL885,wang-chang:2016:P16-1} 
replaced the linear scoring function of each arc in traditional models with neural networks 
and used a margin-based objective~\cite{McDonald:2005} for model training. 
Other work~\cite{zhang2016dependency,dozat2016deep} formalized dependency parsing as independently 
selecting the head of each word with cross-entropy objective, without the guarantee of a 
general non-projective tree structure output. Moreover, there have yet been no previous work on deriving a 
neural probabilistic parsing model to define a proper conditional distribution over non-projective trees
for a given sentence.

In this paper, we propose a probabilistic neural network-based model for non-projective dependency parsing.
This parsing model uses bi-directional LSTM-CNNs (BLSTM-CNNs) as backbone to learn neural 
information representations, on top of which a probabilistic structured layer is constructed with 
a conditional log-linear model, defining a conditional distribution over all non-projective dependency trees. 
The architecture of BLSTM-CNNs is similar to the one used for sequence labeling 
tasks~\cite{ma-hovy:2016:P16-1}, where CNNs encode character-level information of a word into 
its character-level representation and BLSTM models context information of each word. 
Due to the probabilistic structured output layer, we can use negative 
log-likelihood as the training objective, where the partition function and marginals can be computed via 
Kirchhoff's Matrix-Tree Theorem~\cite{tutte1984graph} to process the optimization efficiently by 
back-propagation. At test time, parsing trees can be decoded with the maximum spanning 
tree (MST) algorithm~\cite{McDonald:2005b}. We evaluate our model on 17 treebanks across 14 
different languages, achieving state-of-the-art performance on 9 treebanks. 
The contributions of this work are summarized as: (i) proposing a neural probabilistic model for non-projective 
dependency parsing. (ii) giving empirical evaluations of this model on benchmark data sets over 14 languages. 
(iii) achieving state-of-the-art performance with this parser on nine different treebanks.

\section{Neural Probabilistic Parsing Model}
In this section, we describe the components (layers) of our neural parsing model.
We introduce the neural layers in our neural network one-by-one from top to bottom.

\subsection{Edge-Factored Parsing Layer}
\label{subsec:parser}
\begin{figure}[t]
\epsfig{figure=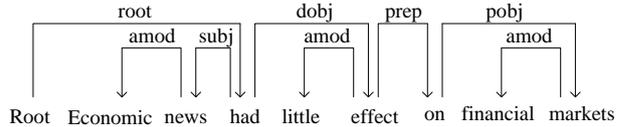,width=0.5\textwidth}
\caption{An example labeled dependency tree.}
\label{fig:dp-tree}
\end{figure}

In this paper, we will use the following notation: $\mathbf{x} = \{x_1, \ldots, x_n \}$ represents 
a generic input sentence, where $x_i$ is the $i$th word.
$\mathbf{y}$ represents a generic (possibly non-projective) dependency tree, which represents syntactic relationships 
through labeled directed edges between heads and their dependents. 
For example, Figure~\ref{fig:dp-tree} shows a dependency tree for the sentence, ``Economic news
had little effect on financial markets'', with the sentence’s root-symbol as its root.
$T(\mathbf{x})$ is used to denote the set of possible dependency trees for sentence $\mathbf{x}$.

The probabilistic model for dependency parsing defines a family of conditional probability 
$p(\mathbf{y}|\mathbf{x}; \Theta)$ over all $\mathbf{y}$ given sentence $\mathbf{x}$,
with a log-linear form:
\begin{displaymath}
P(\mathbf{y}|\mathbf{x}; \Theta) = \frac{\exp\left(\sum\limits_{(x_h, x_m) \in \mathbf{y}} \phi(x_h, x_m; \Theta)\right)}{Z(\mathbf{x}; \Theta)}
\end{displaymath}
where $\Theta$ is the parameter of this model, $s_{hm} = \phi(x_h, x_m; \Theta)$ is 
the score function of edge from $x_h$ to $x_m$, and
\begin{displaymath}
Z(\mathbf{x}; \Theta) = \sum\limits_{\mathbf{y} \in T(\mathbf{x})} \exp\left(\sum\limits_{(x_h, x_m) \in \mathbf{y}} s_{hm} \right)
\end{displaymath}
is the partition function.
\paragraph{Bi-Linear Score Function.} In our model, we adopt a bi-linear form score function: 
{\setlength\arraycolsep{2pt}
\begin{displaymath}
\begin{array}{rcl}
\phi(x_h, x_m; \Theta) & = & \varphi(x_h)^T \mathbf{W} \varphi(x_m) \\
 & & + \mathbf{U}^T\varphi(x_h) + \mathbf{V}^T\varphi(x_m) + \mathbf{b}
\end{array}
\end{displaymath}}
where $\Theta= \{\mathbf{W}, \mathbf{U}, \mathbf{V}, \mathbf{b}\}$, 
$\varphi(x_i)$ is the representation vector of $x_i$, $\mathbf{W}, \mathbf{U}, \mathbf{V}$ denote the 
weight matrix of the bi-linear term and the two weight vectors of the linear terms in $\phi$, 
and $\mathbf{b}$ denotes the bias vector. 

As discussed in \newcite{dozat2016deep}, the bi-linear form of 
score function is related to the bi-linear attention mechanism~\cite{luong-pham-manning:2015:EMNLP}.
The bi-linear score function differs from the traditional score function proposed in \newcite{TACL885} 
by adding the bi-linear term. A similar score function is proposed in \newcite{dozat2016deep}. 
The difference between their and our score function is that they only used the linear term 
for head words ($\mathbf{U}^T\varphi(x_h)$) while use them for both heads and modifiers.

\paragraph{Matrix-Tree Theorem.}
In order to train the probabilistic parsing model, as discussed in \newcite{koo-EtAl:2007:EMNLP-CoNLL2007},
we have to compute the \emph{partition function} and the \emph{marginals}, requiring summation over 
the set $T(\mathbf{x})$:
{\setlength\arraycolsep{2pt}
\begin{displaymath}
\begin{array}{rcl}
Z(\mathbf{x}; \Theta) & = &  \sum\limits_{\mathbf{y} \in T(\mathbf{x})} \prod\limits_{(x_h, x_m) \in \mathbf{y}}\psi(x_h, x_m; \Theta) \\
\mu_{h,m}(\mathbf{x}; \Theta) & = & \sum\limits_{\mathbf{y} \in T(\mathbf{x}):(x_h, x_m) \in \mathbf{y}} P(\mathbf{y}|\mathbf{x}; \Theta)
\end{array}
\end{displaymath}}
where $\psi(x_h, x_m; \Theta)$ is the potential function:
\begin{displaymath}
\psi(x_h, x_m; \Theta) = \exp \left( \phi(x_h, x_m; \Theta)\right)
\end{displaymath}
and 
$\mu_{h,m}(\mathbf{x}; \Theta)$ is the marginal for edge from $h$th word to $m$th word for $\mathbf{x}$.

Previous studies~\cite{koo-EtAl:2007:EMNLP-CoNLL2007,smith-smith:2007} have presented how a variant of 
Kirchhoff's Matrix-Tree Theorem~\cite{tutte1984graph} can be used to evaluate the partition function and
marginals efficiently. In this section, we briefly revisit this method.

For a sentence $\mathbf{x}$ with $n$ words, we denote $\mathbf{x} = \{x_0, x_1, \ldots, x_n\}$, where $x_0$ 
is the root-symbol. We define a complete graph $G$ on $n + 1$ nodes (including the root-symbol $x_0$), 
where each node corresponds to a word in $\mathbf{x}$ and each edge corresponds to a dependency arc 
between two words. Then, we assign non-negative weights to the edges of this complete graph with $n+1$ nodes, 
yielding the weighted adjacency 
matrix $\mathbf{A}(\Theta) \in \mathbb{R}^{n+1 \times n+1}$, for $h, m = 0, \ldots, n$:
\begin{displaymath}
\mathbf{A}_{h,m}(\Theta) = \psi(x_h, x_m; \Theta)
\end{displaymath}
Based on the adjacency matrix $\mathbf{A}(\Theta)$, we have the Laplacian matrix:
\begin{displaymath}
\mathbf{L}(\Theta) = \mathbf{D}(\Theta) - \mathbf{A}(\Theta)
\end{displaymath}
where $\mathbf{D}(\Theta)$ is the weighted degree matrix:
\begin{displaymath}
\mathbf{D}_{h,m}(\Theta) = \left\{
\begin{array}{ll}
\sum\limits_{h'=0}^{n} \mathbf{A}_{h',m}(\Theta) & \textrm{if } h = m \\
0 & \textrm{otherwise}
\end{array}\right.
\end{displaymath}
Then, according to Theorem 1 in \newcite{koo-EtAl:2007:EMNLP-CoNLL2007}, the partition function is equal to
the minor of $\mathbf{L}(\Theta)$ w.r.t row $0$ and column $0$:
\begin{displaymath}
Z(\mathbf{x}; \Theta) = \mathbf{L}^{(0,0)}(\Theta)
\end{displaymath}
where for a matrix $\mathbf{A}$, $\mathbf{A}^{(h, m)}$ denotes the \emph{minor} of 
$\mathbf{A}$ w.r.t row $h$ and column $m$; i.e., the determinant of the submatrix formed by deleting the 
$h$th row and $m$th column.

The marginals can be computed by calculating the matrix inversion of the matrix 
corresponding to $\mathbf{L}^{(0,0)}(\Theta)$. The time complexity of computing the partition function and 
marginals is $O(n^3)$.

\paragraph{Labeled Parsing Model.} Though it is originally designed for unlabeled parsing, 
our probabilistic parsing model is easily extended to include dependency labels.

In labeled dependency trees, each edge is represented by a tuple $(x_h, x_m, l)$,
where $x_h$ and $x_m$ are the head word and modifier, respectively, and $l$ is the label of dependency 
type of this edge. Then we can extend the original model for labeled dependency parsing 
by extending the score function to include dependency labels:
\begin{displaymath}
\begin{array}{rcl}
\phi(x_h, x_m, l; \Theta) & = & \varphi(x_h)^T \mathbf{W}_l \varphi(x_m) \\
 & & + \mathbf{U}_{l}^T\varphi(x_h) + \mathbf{V}_{l}^T\varphi(x_m) \\
 & & + \mathbf{b}_{l}
\end{array}
\end{displaymath}
where $\mathbf{W}_l, \mathbf{U}_l, \mathbf{V}_l, \mathbf{b}_l$ are the weights and bias corresponding 
to dependency label $l$. Suppose that there are $L$ different dependency labels, it suffices to define the
new adjacency matrix by assigning the weight of a edge with the sum of weights over different dependency labels:
\begin{displaymath}
\mathbf{A'}_{h,m}(\Theta) = \sum\limits_{l=1}^{L}\psi(x_h, x_m, l; \Theta)
\end{displaymath}
The partition function and marginals over labeled dependency trees are obtained by operating on the new
adjacency matrix $\mathbf{A'}(\Theta)$. The time complexity becomes $O(n^3 + Ln^2)$. 
In practice, $L$ is probably large. For English, the number of edge labels in 
Stanford Basic Dependencies~\citep{de2006generating} is 45, 
and the number in the treebank of CoNLL-2008 shared task~\citep{surdeanu2008conll} is 70.
While, the average length of sentences in English Penn Treebank~\citep{Marcus:1993} is around 23.
Thus, $L$ is not negligible comparing to $n$.

It should be noticed that in our labeled model, for different dependency label $l$ we use the same 
vector representation $\varphi(x_i)$ for each word $x_i$. The dependency labels are distinguished (only) by
the parameters (weights and bias) corresponding to each of them. One advantage of this is that it 
significantly reduces the memory requirement comparing to the model in \newcite{dozat2016deep} which 
distinguishes $\varphi_{l}(x_i)$ for different label $l$.

\paragraph{Maximum Spanning Tree Decoding.} The decoding problem of this parsing model can be formulated as:
\begin{displaymath}
\begin{array}{rl}
\mathbf{y}^{*} & = \argmax\limits_{\mathbf{y} \in T(\mathbf{x})} P(\mathbf{y}|\mathbf{x}; \Theta) \\
 & = \argmax\limits_{\mathbf{y} \in T(\mathbf{x})} \sum\limits_{(x_h, x_m) \in \mathbf{y}} \phi(x_h, x_m; \Theta)
\end{array}
\end{displaymath}
which can be solved by using the Maximum Spanning Tree (MST) algorithm described in \newcite{McDonald:2005b}.

\subsection{Neural Network for Representation Learning}
Now, the remaining question is how to obtain the vector representation of each word with a neural network.
In the following subsections, we will describe the architecture of our neural network model for representation 
learning.

\begin{figure}[t]
\centering
\includegraphics[scale=0.72]{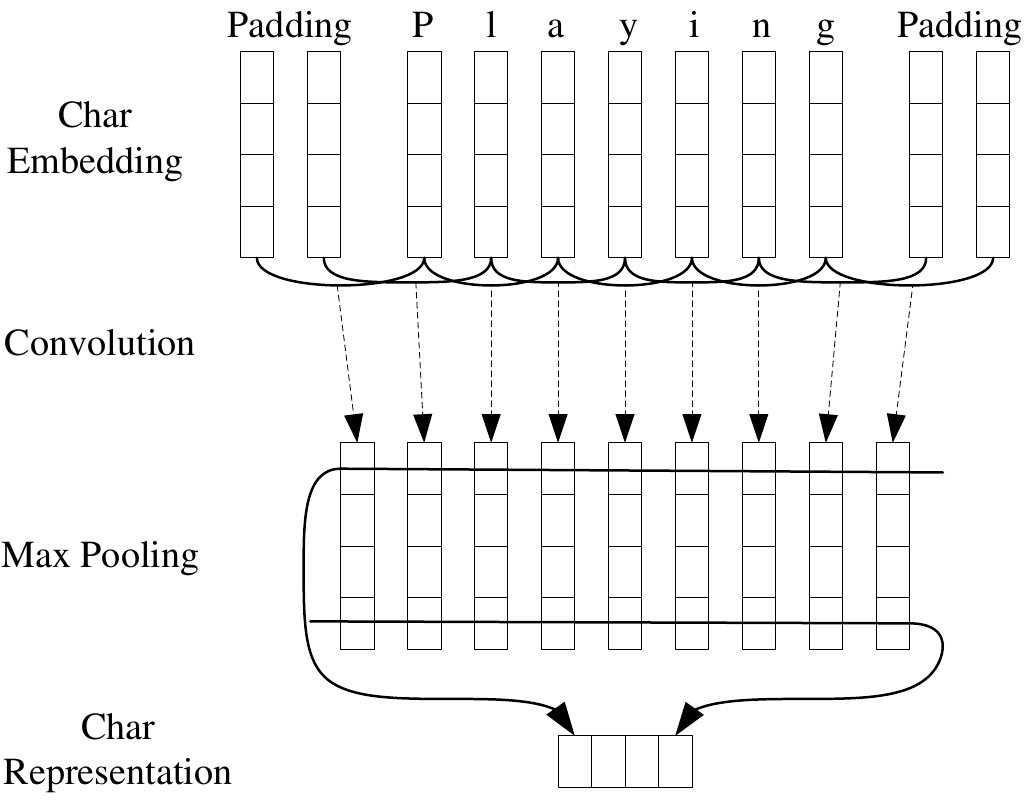}
\caption{The convolution neural network for extracting character-level representations of words. Dashed arrows indicate a dropout layer applied before character embeddings are input to CNN.}
\label{fig:cnn}
\end{figure}

\subsubsection{CNNs}
Previous work~\cite{santos2014learning} have shown that CNNs are an 
effective approach to extract morphological information (like the prefix or suffix of a word) 
from characters of words and encode it into neural representations, which has been proven 
particularly useful on Out-of-Vocabulary words (OOV). 
The CNN architecture our model uses to extract character-level representation 
of a given word is the same as the one used in ~\newcite{ma-hovy:2016:P16-1}. 
The CNN architecture is shown in Figure~\ref{fig:cnn}. Following~\newcite{ma-hovy:2016:P16-1}, 
a dropout layer~\cite{srivastava2014dropout} is applied before character embeddings are input to CNN.

\subsubsection{Bi-directional LSTM}
\paragraph{LSTM Unit.} Recurrent neural networks (RNNs) are a powerful family of connectionist models 
that have been widely applied in NLP tasks, such as language modeling~\cite{mikolov2010recurrent}, 
sequence labeling~\cite{ma-hovy:2016:P16-1} and machine translation~\cite{cho2014properties}, 
to capture context information in languages. 
Though, in theory, RNNs are able to learn long-distance dependencies, 
in practice, they fail due to the gradient vanishing/exploding problems~\cite{bengio1994,pascanu2013}.

LSTMs~\cite{hochreiter1997long} are variants of RNNs designed to cope with these gradient vanishing problems.
Basically, a LSTM unit is composed of three multiplicative gates which control the proportions of 
information to pass and to forget on to the next time step.

\paragraph{BLSTM.}
Many linguistic structure prediction tasks can benefit from having access to both past (left) 
and future (right) contexts, while the LSTM's hidden state $\mathbf{h}_t$ takes information only from past, 
knowing nothing about the future. 
An elegant solution whose effectiveness has been proven by previous 
work~\cite{dyer-EtAl:2015:ACL-IJCNLP,ma-hovy:2016:P16-1} is bi-directional LSTM (BLSTM). 
The basic idea is to present each sequence forwards and backwards to two separate hidden states to 
capture past and future information, respectively. 
Then the two hidden states are concatenated to form the final output.
As discussed in \newcite{dozat2016deep}, there are more than one advantages to apply a multilayer 
perceptron (MLP) to the output vectors of BLSTM before the score function, eg. reducing the dimensionality
and overfitting of the model. We follow this work by using a one-layer perceptron with 
elu~\cite{clevert2015fast} as activation function.

\subsection{BLSTM-CNNs}
\begin{figure}[t]
\centering
\includegraphics[scale=0.9]{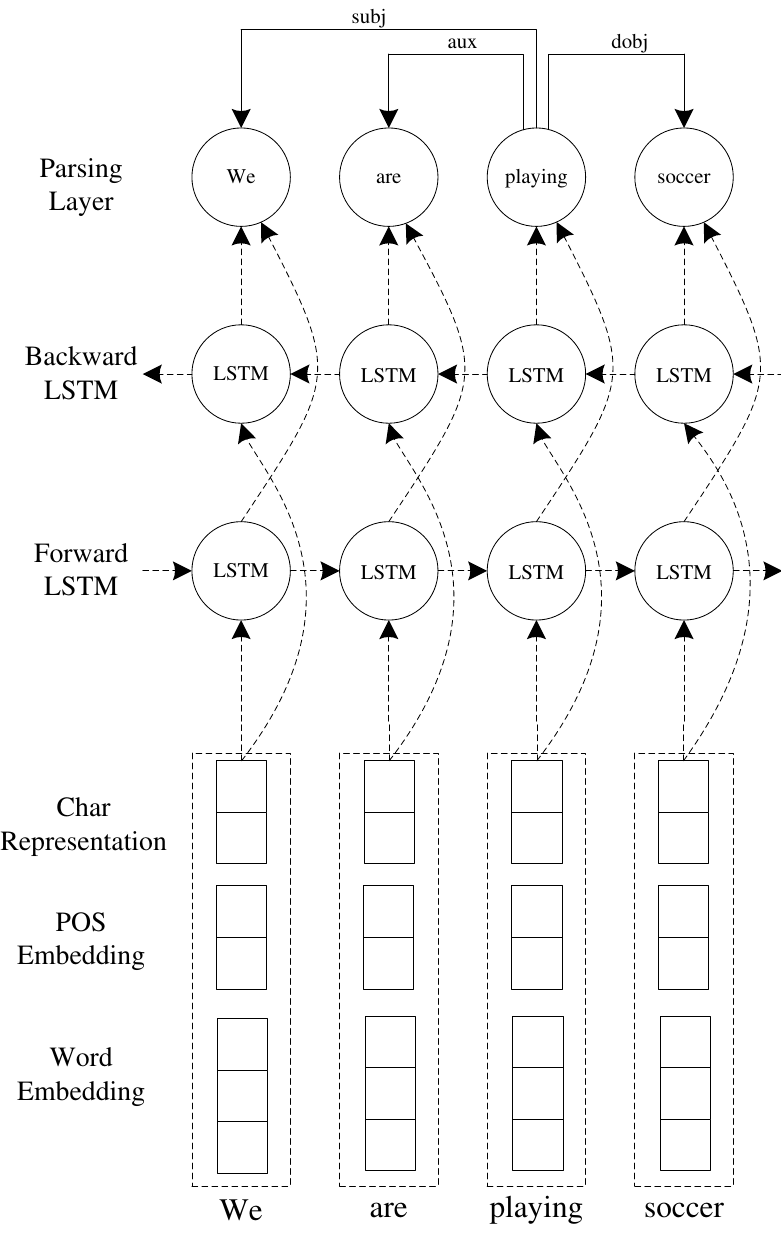}
\caption{The main architecture of our parsing model. The character representation for each word is 
computed by the CNN in Figure~\ref{fig:cnn}. Then the character representation vector is concatenated 
with the word and pos embedding before feeding into the BLSTM network. 
Dashed arrows indicate dropout layers applied on the input, hidden and output vectors of BLSTM.}
\label{fig:architec}
\end{figure}

Finally, we construct our neural network model by feeding the output vectors of BLSTM
(after MLP) into the parsing layer. 
Figure~\ref{fig:architec} illustrates the architecture of our network in detail.

For each word, the CNN in Figure~\ref{fig:cnn}, with character embeddings as inputs, encodes 
the character-level representation. Then the character-level representation vector is concatenated 
with the word embedding vector to feed into the BLSTM network. To enrich word-level information, 
we also use POS embeddings. Finally, the output vectors of the neural netwok are fed to the
parsing layer to jointly parse the best (labeled) dependency tree. 
As shown in Figure~\ref{fig:architec}, dropout layers are applied on the input, hidden and output 
vectors of BLSTM, using the form of recurrent dropout proposed in~\newcite{gal2016dropout:rnn}.

\section{Network Training}
In this section, we provide details about implementing and training the neural parsing model, 
including parameter initialization, model optimization and hyper parameter selection.

\begin{table}
\centering
\begin{tabular}[t]{l|l|r}
\hline
\textbf{Layer} & \textbf{Hyper-parameter} & \textbf{Value} \\
\hline
\multirow{2}{*}{CNN} & window size & 3 \\
 & number of filters & 50 \\
\hline
\multirow{4}{*}{LSTM} & number of layers & 2 \\
 & state size & 256\\
 & initial state & 0.0 \\
 & peepholes & Hadamard\\
\hline
\multirow{2}{*}{MLP} & number of layers & 1 \\
 & dimension & 100 \\
\hline
\multirow{3}{*}{Dropout} & embeddings & 0.15 \\
 & LSTM hidden states & 0.25 \\
 & LSTM layers & 0.33 \\
\hline
\multirow{4}{*}{Learning} & optimizer & Adam \\
 & initial learning rate & 0.002 \\
 & decay rate & 0.5 \\
 & gradient clipping & 5.0 \\
\hline
\end{tabular}
\caption{Hyper-parameters for all experiments.}
\label{tab:hyper-params}
\end{table}

\begin{table*}
\centering
{\small
\begin{tabular}[t]{l|cc:cc|cc:cc|cc:cc}
\hline
 & \multicolumn{4}{c|}{\textbf{English}} & \multicolumn{4}{c|}{\textbf{Chinese}} & \multicolumn{4}{c}{\textbf{German}} \\
 \cline{2-13}
 & \multicolumn{2}{c:}{\textbf{Dev}} & \multicolumn{2}{c|}{\textbf{Test}} & \multicolumn{2}{c:}{\textbf{Dev}} & \multicolumn{2}{c|}{\textbf{Test}} & \multicolumn{2}{c:}{\textbf{Dev}} & \multicolumn{2}{c}{\textbf{Test}} \\
 \cline{2-13}
\textbf{Model} & UAS & LAS & UAS & LAS & UAS & LAS & UAS & LAS & UAS & LAS & UAS & LAS \\
\hline
\textsf{Basic} & 94.51 & 92.23 & 94.62 & 92.54 & 84.33 & 81.65 & 84.35 & 81.63 & 90.46 & 87.77 & 90.69 & 88.42 \\
\textsf{+Char} & 94.74 & 92.55 & 94.73 & 92.75 & 85.07 & 82.63 & 85.24 & 82.46 & 92.16 & 89.82 & 92.24 & 90.18 \\
\textsf{+POS} & 94.71 & 92.60 & 94.83 & 92.96 & 88.98 & 87.55 & 89.05 & 87.74 & 91.94 & 89.51 & 92.19 & 90.05 \\
\textsf{Full} & 94.77 & 92.66 & 94.88 & 92.98 & 88.51 & 87.16 & 88.79 & 87.47 & 92.37 & 90.09 & 92.58 & 90.54 \\
\hline
\end{tabular}
}
\caption{Parsing performance (UAS and LAS) of different versions of our model on 
both the development and test sets for three languages.}
\label{tab:main:results}
\end{table*}

\subsection{Parameter Initialization}
\paragraph{Word Embeddings.} For all the parsing models on different languages, 
we initialize word vectors with pretrained word embeddings. For Chinese, Dutch, English, German and Spanish,
we use the structured-skipgram~\cite{ling-EtAl:2015:NAACL-HLT} embeddings, and for other languages we use 
the Polyglot~\cite{polyglot:2013:CoNLL} embeddings. The dimensions of embeddings are 100 for English, 50 for
Chinese and 64 for other languages. 

\paragraph{Character Embeddings.} 
Following \newcite{ma-hovy:2016:P16-1}, character embeddings are initialized with uniform samples from 
$[-\sqrt{\frac{3}{dim}}, +\sqrt{\frac{3}{dim}}]$, where we set $dim=50$.

\paragraph{POS Embedding.} Our model also includes POS embeddings. The same as character embeddings, 
POS embeddings are also 50-dimensional, initialized uniformly 
from $[-\sqrt{\frac{3}{dim}}, +\sqrt{\frac{3}{dim}}]$.

\paragraph{Weights Matrices and Bias Vectors.} Matrix parameters are randomly initialized with uniform 
samples from $[-\sqrt{\frac{6}{r + c}}, +\sqrt{\frac{6}{r + c}}]$, where $r$ and $c$ are the number of 
of rows and columns in the structure~\cite{glorot2010}. Bias vectors are initialized to zero, 
except the bias $\mathbf{b}_f$ for the forget gate in LSTM , which is initialized to 1.0~\cite{jozefowicz2015}.

\subsection{Optimization Algorithm}
Parameter optimization is performed with the Adam optimizer~\cite{kingma2014adam} with $\beta1 = \beta2 = 0.9$.
We choose an initial learning rate of $\eta_0 = 0.002$. The learning rate $\eta$ was adapted using a 
schedule $S = [e_1, e_2, \ldots, e_s]$, in which the learning rate $\eta$ is annealed by multiplying a fixed
decay rate $\rho = 0.5$ after $e_i \in S$ epochs respectively. We used $S = [10,30,50, 70, 100]$ and 
trained all networks for a total of 120 epochs. While the Adam optimizer automatically adjusts the global 
learning rate according to past gradient magnitudes, we find that this additional decay consistently improves
model performance across all settings and languages.
To reduce the effects of ``gradient exploding'', we use a gradient clipping of $5.0$~\cite{pascanu2013}.
We explored other optimization algorithms such as stochastic gradient descent (SGD) with momentum, 
AdaDelta~\cite{zeiler2012adadelta}, or RMSProp~\cite{dauphin2015rmsprop}, but none of them meaningfully 
improve upon Adam with learning rate annealing in our preliminary experiments. 

\paragraph{Dropout Training.} To mitigate overfitting, we apply the dropout method~\cite{srivastava2014dropout,iclr2017:ma:dropout} 
to regularize our model. As shown in Figure~\ref{fig:cnn} and \ref{fig:architec}, 
we apply dropout on character embeddings before inputting to CNN, and on the input, hidden and output vectors 
of BLSTM. We apply dropout rate of 0.15 to all the embeddings. 
For BLSTM, we use the recurrent dropout~\cite{gal2016dropout:rnn} with 0.25 dropout rate between hidden states
and 0.33 between layers. We found that the model using the new recurrent dropout converged much faster than standard dropout, while achiving similar performance.

\subsection{Hyper-Parameter Selection}
Table~\ref{tab:hyper-params} summarizes the chosen hyper-parameters for all experiments. 
We tune the hyper-parameters on the development sets by random search. 
We use the same hyper-parameters across the models on different treebanks and languages, due to time constrains.
Note that we use 2-layer BLSTM followed with 1-layer MLP. We set the state size of LSTM to $256$ and the 
dimension of MLP to $100$.
Tuning these two parameters did not significantly impact the performance of our model.
\begin{table}
\centering
{\small
\begin{tabular}[t]{l|cc:cc}
\hline
 & \multicolumn{2}{c:}{\textbf{Dev}} & \multicolumn{2}{c}{\textbf{Test}} \\
\hline
 & UAS & LAS & UAS & LAS \\
\hline
cross-entropy & 94.10 & 91.52 & 93.77 & 91.57 \\
global-likelihood & 94.77 & 92.66 & 94.88 & 92.98 \\
\hline
\end{tabular}
}
\caption{Parsing performance on PTB with different training objective functions.}
\label{tab:objective}
\end{table}

\section{Experiments}
\label{sec:experiment}
\subsection{Setup}
We evaluate our neural probabilistic parser on the same data setup as \newcite{kuncoro-EtAl:2016:EMNLP2016}, 
namely the English Penn Treebank (PTB version 3.0)~\citep{Marcus:1993}, 
the Penn Chinese Treebank (CTB version 5.1)~\cite{xue2002building}, 
and the German CoNLL 2009 corpus~\cite{hajivc2009conll}.
Following previous work, all experiments are evaluated on the metrics of unlabeled attachment score (UAS) 
and Labeled attachment score (LAS).

\begin{table*}
\centering
\begin{tabular}[t]{l|cc|cc|cc}
\hline
& \multicolumn{2}{c|}{\textbf{English}} & \multicolumn{2}{c|}{\textbf{Chinese}} & \multicolumn{2}{c}{\textbf{German}} \\
\cline{2-7}
\textbf{System} & UAS & LAS & UAS & LAS & UAS & LAS \\
\hline
\newcite{bohnet-nivre:2012:EMNLP-CoNLL} & -- & -- & 87.3 & 85.9 & 91.4 & 89.4 \\
\newcite{chen-manning:2014:EMNLP2014} & 91.8 & 89.6 & 83.9 & 82.4 & -- & -- \\
\newcite{ballesteros-dyer-smith:2015:EMNLP} & 91.6 & 89.4 & 85.3 & 83.7 & 88.8 & 86.1 \\
\newcite{dyer-EtAl:2015:ACL-IJCNLP} & 93.1 & 90.9 & 87.2 & 85.7 & -- & -- \\
\newcite{TACL885}: graph & 93.1 & 91.0 & 86.6 & 85.1 & -- & -- \\
\newcite{ballesteros-EtAl:2016:EMNLP2016} & 93.6 & 91.4 & 87.7 & 86.2 & -- & -- \\
\newcite{wang-chang:2016:P16-1} & 94.1 & 91.8 & 87.6 & 86.2 & -- & -- \\
\newcite{zhang2016dependency} & 94.1 & 91.9 & 87.8 & 86.2 & -- & -- \\
\newcite{cheng-EtAl:2016:EMNLP2016} & 94.1 & 91.5 & 88.1 & 85.7 & -- & -- \\
\newcite{andor-EtAl:2016:P16-1} & 94.6 & 92.8 & -- & -- & 90.9 & 89.2 \\
\newcite{kuncoro-EtAl:2016:EMNLP2016} & 94.3 & 92.1 & 88.9 & 87.3 & 91.6 & 89.2 \\
\newcite{dozat2016deep} & \textbf{95.7} & \textbf{94.1} & \textbf{89.3} & \textbf{88.2} & \textbf{93.5} & \textbf{91.4} \\
\hline
This work: \textsf{Basic} & 94.6 & 92.5 & 84.4 & 81.6 & 90.7 & 88.4 \\
This work: \textsf{+Char} & 94.7 & 92.8 & 85.2 & 82.5 & 92.2 & 90.2 \\
This work: \textsf{+POS}  & 94.8 & 93.0 & 89.1 & 87.7 & 92.2 & 90.1 \\
This work: \textsf{Full}  & 94.9 & 93.0 & 88.8 & 87.5 & 92.6 & 90.5 \\
\hline
\end{tabular}
\caption{UAS and LAS of four versions of our model on test sets for three languages, 
together with top-performance parsing systems.}
\label{tab:main:comparison}
\end{table*}

\subsection{Main Results}
\label{subsec:main:results}
We first construct experiments to dissect the effectiveness of each input information (embeddings) 
of our neural network architecture by ablation studies. We compare the performance of four versions of our 
model with different inputs --- \textsf{Basic}, \textsf{+POS}, \textsf{+Char} and \textsf{Full} --- where 
the \textsf{Basic} model utilizes only the pretrained word embeddings as inputs, while the \textsf{+POS}
and \textsf{+Char} models augments the basic one with POS embedding and character information, respectively.
According to the results shown in Table~\ref{tab:main:results}, \textsf{+Char} model obtains better performance
than the \textsf{Basic} model on all the three languages, showing that character-level representations 
are important for dependency parsing. Second, on English and German, \textsf{+Char} and \textsf{+POS} achieves
comparable performance, while on Chinese \textsf{+POS} significantly outperforms \textsf{+Char} model.
Finally, 
the \textsf{Full} model achieves the best accuracy 
on English and German, but on Chinese \textsf{+POS} obtains the best. Thus, we guess
that the POS information is more useful for Chinese than English and German.

Table~\ref{tab:objective} gives the performance on PTB of the parsers trained with two different objective 
functions --- the cross-entropy objective of each word, and our objective based on likelihood for an entire tree. 
The parser with global likelihood objective outperforms the one with simple cross-entropy objective, 
demonstrating the effectiveness of the global structured objective.

\subsection{Comparison with Previous Work}
Table~\ref{tab:main:comparison} illustrates the results of the four versions of our model on the three 
languages, together with twelve previous top-performance systems for comparison. Our \textsf{Full} model
significantly outperforms the graph-based parser proposed in \newcite{TACL885} which used similar neural
network architecture for representation learning (detailed discussion in Section~\ref{sec:relatedwork}).
Moreover, our model achieves better results than the parser distillation 
method~\cite{kuncoro-EtAl:2016:EMNLP2016} on all the three languages.
The results of our parser are slightly worse than the scores reported in \newcite{dozat2016deep}. 
One possible reason is that, as mentioned in Section~\ref{subsec:parser}, 
for labeled dependency parsing \newcite{dozat2016deep} 
used different vectors for different dependency labels to represent each word, 
making their model require much more memory than ours.

\begin{table*}
\centering
{\small
\begin{tabular}[t]{l|c:c:c:c:c|c:c||cc}
\hline
 & \textbf{Turbo} & \textbf{Tensor} & \textbf{RGB} & \textbf{In-Out} & \textbf{Bi-Att} & 
\textbf{\textsf{+POS}} & \textbf{\textsf{Full}} & \multicolumn{2}{c}{\textbf{Best Published}} \\
\cline{2-10}
 & UAS & UAS & UAS & UAS [LAS] & UAS [LAS] & UAS [LAS] & UAS [LAS] & UAS & LAS \\
\hline
ar & 79.64 & 79.95 & 80.24 & 79.60 [67.09] & 80.34 [68.58] & 80.05 [67.80] & 80.80 [\textbf{69.40}] & \textbf{81.12} & -- \\
bg & 93.10 & 93.50 & 93.72 & 92.68 [87.79] & 93.96 [89.55] & 93.66 [89.79] & \textbf{94.28} [\textbf{90.60}] & 94.02 & -- \\
zh & 89.98 & 92.68 & 93.04 & 92.58 [88.51] & --  & \textbf{93.44} [90.04] & 93.40 [\textbf{90.10}] &  93.04 & -- \\
cs & 90.32 & 90.50 & 90.77 & 88.01 [79.31] & 91.16 [85.14] & 91.04 [85.82] & \textbf{91.18} [\textbf{85.92}] & 91.16 & 85.14 \\
da & 91.48 & 91.39 & 91.86 & 91.44 [85.55] & 91.56 [85.53] & 91.52 [86.57] & 91.86 [\textbf{87.07}] & \textbf{92.00} & -- \\
nl & 86.19 & 86.41 & 87.39 & 84.45 [80.31] & 87.15 [82.41] & 87.41 [84.17] & \textbf{87.85} [\textbf{84.82}] & 87.39 & -- \\
en & 93.22 & 93.02 & 93.25 & 92.45 [89.43] & -- & 94.43 [92.31] & \textbf{94.66} [\textbf{92.52}] &  93.25 & -- \\
de & 92.41 & 91.97 & 92.67 & 90.79 [87.74] & 92.71 [89.80] & 93.53 [91.55] & \textbf{93.62} [\textbf{91.90}] & 92.71 & 89.80 \\
ja & 93.52 & 93.71 & 93.56 & 93.54 [91.80] & 93.44 [90.67] & 93.82 [92.34] & \textbf{94.02} [\textbf{92.60}] & 93.80 & -- \\
pt & 92.69 & 91.92 & 92.36 & 91.54 [87.68] & 92.77 [88.44] & 92.59 [\textbf{89.12}] & 92.71 [88.92] & \textbf{93.03} & -- \\
sl & 86.01 & 86.24 & 86.72 & 84.39 [73.74] & 86.01 [75.90] & 85.73 [76.48] & 86.73 [\textbf{77.56}] & \textbf{87.06} & -- \\
es & 85.59 & 88.00 & 88.75 & 86.44 [83.29] & 88.74 [84.03] & 88.58 [85.03] & \textbf{89.20} [\textbf{85.77}] & 88.75 & 84.03 \\
sv & 91.14 & 91.00 & 91.08 & 89.94 [83.09] & 90.50 [84.05] & 90.89 [86.58] & 91.22 [\textbf{86.92}] & \textbf{91.85} & 85.26 \\
tr & 76.90 & 76.84 & 76.68 & 75.32 [60.39] & \textbf{78.43} [\textbf{66.16}] & 75.88 [61.72] & 77.71 [65.81] & 78.43 & 66.16 \\
\hline
av & 88.73 & 89.08 & 89.44 & 88.08 [81.84] & -- & 89.47 [84.24] & 89.95 [84.99] & 89.83 & -- \\
\end{tabular}
}
\caption{UAS and LAS on 14 treebanks from CoNLL shared tasks, together with several state-of-the-art parsers.
``Best Published'' includes the most accurate parsers in term of UAS among \newcite{koo-EtAl:2010:EMNLP}, 
\newcite{martins-EtAl:2011:EMNLP1}, \newcite{martins:2013:Short}, \newcite{lei-EtAl:2014}, 
\newcite{zhang-EtAl:2014:EMNLP20143},  \newcite{zhang-mcdonald:2014}, \newcite{pitler-mcdonald:2015},
\newcite{ma-hovy:2015:EMNLP}, and \newcite{cheng-EtAl:2016:EMNLP2016}.}
\label{tab:conll}
\end{table*}

\subsection{Experiments on CoNLL Treebanks}
\paragraph{Datasets.} To make a thorough empirical comparison with previous studies, 
we also evaluate our system on treebanks 
from CoNLL shared task on dependency parsing --- the English treebank from CoNLL-2008 
shared task~\cite{surdeanu2008conll} and all 13 treebanks from CoNLL-2006 shared task~\cite{Buchholz:2006}.
For the treebanks from CoNLL-2006 shared task, following \newcite{cheng-EtAl:2016:EMNLP2016}, we randomly 
select 5\% of the training data as the development set. UAS and LAS are evaluated using the official
scorer\footnote{\url{http://ilk.uvt.nl/conll/software.html}} of CoNLL-2006 shared task.

\paragraph{Baselines.} We compare our model with the third-order Turbo parser~\cite{martins:2013:Short}, 
the low-rank tensor based model (Tensor)~\cite{lei-EtAl:2014}, 
the randomized greedy inference based (RGB) model~\cite{zhang-EtAl:2014:EMNLP20143}, 
the labeled dependency parser with inner-to-outer greedy decoding algorithm (In-Out)~\cite{ma-hovy:2015:EMNLP}, 
and the bi-direction attention based parser (Bi-Att)~\cite{cheng-EtAl:2016:EMNLP2016}. 
We also compare our parser against the best published results for individual languages. 
This comparison includes four additional systems: \newcite{koo-EtAl:2010:EMNLP}, 
\newcite{martins-EtAl:2011:EMNLP1}, \newcite{zhang-mcdonald:2014} and \newcite{pitler-mcdonald:2015}.

\paragraph{Results.} Table~\ref{tab:conll} summarizes the results of our model, along with the state-of-the-art
baselines. On average across 14 languages, our approach significantly outperforms all the baseline systems.
It should be noted that the average UAS of our parser over the 14 languages is
better than that of the ``best published'', which are from different systems that achieved best results for
different languages.

For individual languages, our parser achieves state-of-the-art performance on both UAS and LAS on 
8 languages --- Bulgarian, Chinese, Czech, Dutch, English, German, Japanese and Spanish. 
On Arabic, Danish, Portuguese, Slovene and Swedish, our parser obtains the best LAS.
Another interesting observation is that the \textsf{Full} model outperforms the \textsf{+POS} model on 
13 languages. The only exception is Chinese, which matches the observation in Section~\ref{subsec:main:results}.

\section{Related Work}
\label{sec:relatedwork}
In recent years, several different neural network based models have been proposed 
and successfully applied to dependency parsing. Among these neural models, there are three approaches 
most similar to our model --- the two graph-based parsers with BLSTM feature 
representation~\cite{TACL885,wang-chang:2016:P16-1}, and the neural bi-affine attention 
parser~\cite{dozat2016deep}.

\newcite{TACL885} proposed a graph-based dependency parser which uses BLSTM for word-level representations.
\newcite{wang-chang:2016:P16-1} used a similar model with a way to learn sentence segment embedding 
based on an extra forward LSTM network. Both of these two parsers trained the parsing models by optimizing 
margin-based objectives. There are three main differences between their models and ours. 
First, they only used linear form score function, instead of using the bi-linear term between the 
vectors of heads and modifiers. Second, They did not employ CNNs to model character-level information.
Third, we proposed a probabilistic model over non-projective trees on the top of neural representations, 
while they trained their models with a margin-based objective.
\newcite{dozat2016deep} proposed neural parsing model using bi-affine score function, which is similar to 
the bi-linear form score function in our model. Our model mainly differ from this model by using CNN to 
model character-level information. Moreover, their model formalized dependency parsing as independently 
selecting the head of each word with cross-entropy objective, while our probabilistic parsing model 
jointly encodes and decodes parsing trees for given sentences.

\section{Conclusion}
In this paper, we proposed a neural probabilistic model for non-projective dependency parsing, using the 
BLSTM-CNNs architecture for representation learning.  Experimental results on 17 treebanks across 14 languages 
show that our parser significantly improves the accuracy of both dependency structures (UAS) and edge 
labels (LAS), over several previously state-of-the-art systems.

\section*{Acknowledgements}
This research was supported in part by DARPA grant FA8750-12-2-0342 funded under the DEFT program. Any opinions, findings,
and conclusions or recommendations expressed in this material are those of the authors and do not necessarily reflect the views of DARPA.
\bibliography{ijcnlp2017}
\bibliographystyle{ijcnlp2017}
\end{document}